\documentclass[10pt,twocolumn,letterpaper]{article}

\usepackage{cvpr}      

\usepackage[accsupp]{axessibility}

\usepackage{amsthm}
\usepackage{amsmath}
\usepackage{amssymb}

\usepackage{xcolor}
\usepackage{colortbl}
\usepackage{multirow}
\usepackage{booktabs}
\usepackage{pifont}

\usepackage{adjustbox}
\usepackage{newfloat}
\usepackage{listings}
\usepackage{algorithm}
\usepackage{algorithmic}

\usepackage[pagebackref,breaklinks,colorlinks,allcolors=cvprblue]{hyperref}

\newcommand{\tsb}{\textsubscript}

\definecolor{cvprblue}{rgb}{0.21,0.49,0.74}

\def\paperID{911} 
\def\confName{CVPR}
\def\confYear{2025}

\title{Adversarial Domain Prompt Tuning and Generation for Single Domain Generalization}

\author{Zhipeng Xu$^{1}$,De Cheng$^{1}$\thanks{Corresponding authors.}, Xinyang Jiang$^2$, Nannan Wang$^{1}$, Dongsheng Li$^2$, Xinbo Gao$^3$\\ 
{$^{1}$}Xidian University, 
{$^{2}$}Microsoft Research Asia, 
{$^{3}$}Chongqing University of Posts and Telecommunications\\
{\tt \small xu\_zhipeng@stu.xidian.edu.cn,  \{nnwang,dcheng\}@xidian.edu.cn}\\
{\tt \small \{xinyangjiang, dongsli\}@microsoft.com, gaoxb@cqupt.edu.cn}}

\begin{document}
\maketitle

\begin{abstract}
Single domain generalization (SDG) aims to learn a robust model, which could perform well on many unseen domains while there is only one single domain available for training. 
One of the promising directions for achieving single-domain generalization is to generate out-of-domain (OOD) training data through data augmentation or image generation. Given the rapid advancements in AI-generated content (AIGC), this paper is the first to propose leveraging powerful pre-trained text-to-image (T2I) foundation models to create the training data. However, manually designing textual prompts to generate images for all possible domains is often impractical, and some domain characteristics may be too abstract to describe with words.
To address these challenges, we propose a novel Progressive Adversarial Prompt Tuning (PAPT) framework for pre-trained diffusion models. Instead of relying on static textual domains, our approach learns two sets of abstract prompts as conditions for the diffusion model: one that captures domain-invariant category information and another that models domain-specific styles. 
This adversarial learning mechanism enables the T2I model to generate images in various domain styles while preserving key categorical features.
Extensive experiments demonstrate the effectiveness of the proposed method, achieving superior performances to state-of-the-art single-domain generalization approaches. Code is available in the supplementary materials.
\end{abstract}    
\section{Introduction}

Recent machine learning models ($e.g.$, deep neural network) have achieved remarkable performances on various tasks, under the key assumption that the training and testing data are independently and identically distributed ($i.e.$, $iid$). However, this assumption often does not hold in many real-world scenarios due to the domain shift, especially when testing on out-of-distribution (OOD) or previously unseen datasets. To mitigate the issue of domain shift, various approaches have been proposed, including domain adaptation (DA)~\cite{ganin2016domain, long2013transfer, pan2019transferrable, qu2022bmd, qu2023upcycling, cheng2023efficient} and domain generalization (DG) methods~\cite{IKI, StPR, wang2022generalizing, zhang2022towards, cheng2024disentangled, PADG, RD-MLDG}. These methods typically involve incorporating data from multiple training domains ($e.g.$, multi-domain generalization), or the target domain data. Despite showing encouraging performances on the OOD data, it may not always be applicable due to data acquiring budget or privacy issue. Therefore, we aim to address the more challenging but more useful task, $i.e.$, single domain generalization (SDG).

\begin{figure}
    \centering
    \includegraphics[width=0.47\textwidth]{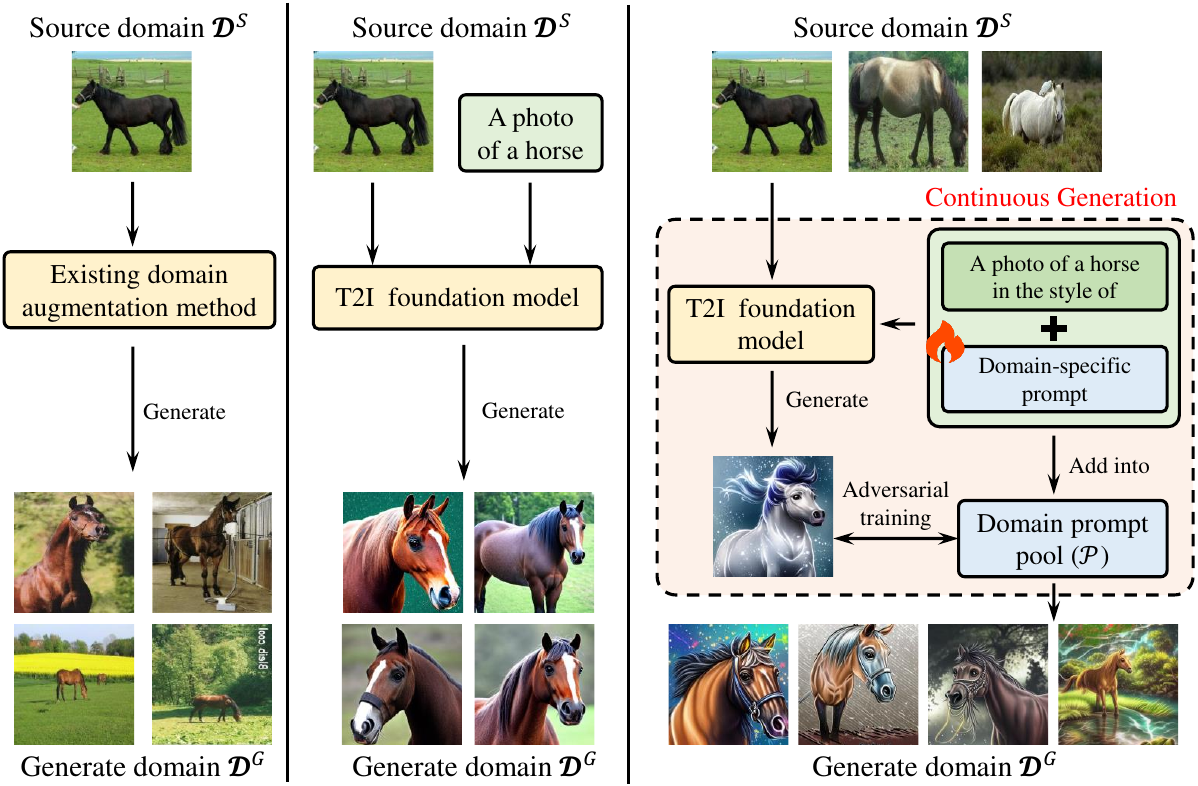}
    \vspace{-2.5mm}
    \caption{
   Comparison between our method with existing single domain generalization approaches: \emph{Left:} the traditional domain augmentation-based method; \emph{Middle:} the simple T2I foundation model with text description as prompt for domain augmentation; \emph{Right:} our proposed method, which learns two sets of abstract prompts ($i.e.$, category and domain) instead of the static textual domains, to instruct the diffusion-based T2I foundation model for generating images of a wide range of domain styles.     
    }
    \label{pic:introduction}
    \vspace{-6.5mm}
\end{figure}

SDG aims to train a model on a labeled source domain and perform well on multiplied unseen target domains.  
This learning paradigm presents significant challenges because only one source domain is given and the target domains are usually unseen/out-of-distribution and unavailable during training.
Existing works have made considerable successes by employing data augmentation to extend the distribution of the source domain~\cite{carlucci2019domain, shankar2018generalizing} or learning adaptive data normalization~\cite{fan2021adversarially} typically.
However, it remains challenging how to construct diverse domains to achieve a provable generalization performance on target domains. 
Namely, it is challenging to guarantee a mitigated distribution discrepancy between the generated domains and target domains.
Recent research provides strong evidence that Text-to-Image (T2I) foundation models have a powerful ability to generate images from text descriptions. Given that text inherently conveys rich semantic information, these models can produce a wide variety of images across different domains using text prompts. 
However, there still contains the following key issues that need to be addressed:
\emph{1) How to generate images from domains whose characteristics that are too abstract to describe with language? 2) How to enable T2I models to automatically generate a sufficiently diverse range of images across different domains?}

To address these challenges, this paper proposes to automatically learn a group of abstract prompts for out-of-distribution image generation with T2I models, eliminating the need for manually crafted descriptions. 
As shown in Fig.~\textcolor{red}{\ref{pic:introduction}}, our method applies prompt-tuning to diffusion-based T2I models, learning two types of abstract prompts: \emph{category prompts}, which capture domain-invariant category information that remains consistent across domains, and \emph{domain prompts}, which represent diverse domain styles, including those too abstract to express in words ($e.g.$, ``a \ photo \ of \ [class\_prompt] \ in \ the \ style \ of \ [domain\_prompt]''). 
Furthermore, to ensure both diversity in domain styles and consistency in key categorical features of generated images, we propose a novel Progressive Adversarial Prompt Tuning (PAPT) framework. In this framework, we first learn category prompts from the source domain data, then progressively learn new domain prompts, which are stored in a domain-specific memory bank. This framework reduces style similarity between newly generated domains and those already in the memory bank while enhancing semantic consistency between generated images and their corresponding category prompts.
The main contributions can be summarized as follows:

\begin{itemize}
\item We propose the first work that leverages the text-to-image foundation model for single-domain generalization (SDG). To obtain a wider range of domain styles beyond manual language description, we propose to learn a group of abstract learnable prompts representing diverse domains instead of static textual domains.
\item 
We propose a novel Progressive Adversarial Prompt Tuning (PAPT) framework to balance the trade-off between generating images in diverse domain styles and preserving key categorical features. 
\item Through generating images with high domain diversity, the proposed method achieves superior performances to state-of-the-art SDG methods, illustrated by extensive experiments on mainstream SDG datasets. 
\end{itemize}

\begin{figure*}[htbp] \label{main_pic}
    \centering
    \includegraphics[width=1.0\textwidth]{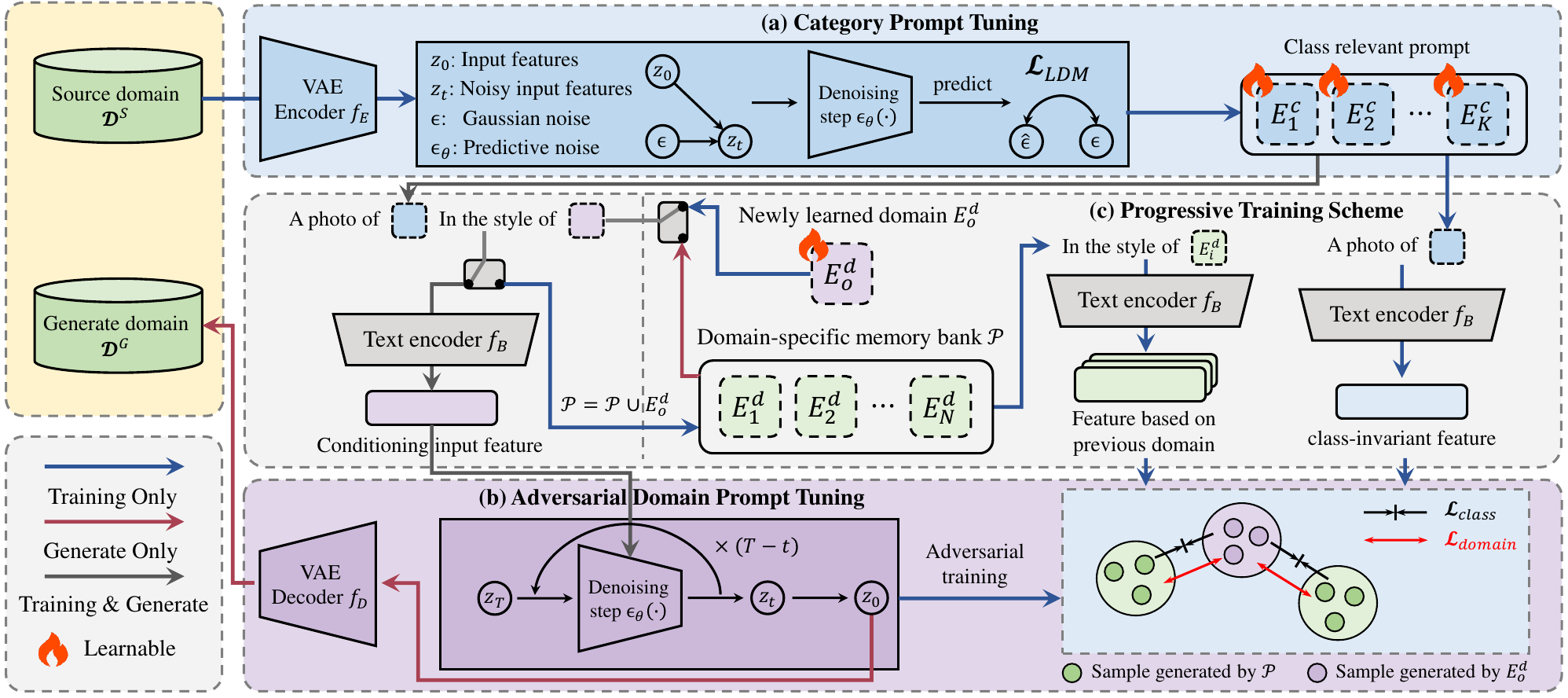}
    \vspace{-6.0mm}
    \caption{
    The framework of our proposed method. 
    We first train a set of category prompts $\mathbf{E}_k^c$ for each class to capture domain-invariant category information in step \textcolor{red}{(a)}.
    Subsequently, we train domain prompts $\mathbf{E}_o^d$ that represent abstract domain styles instead of rigid textual domains through adversarial domain prompt tuning in step \textcolor{red}{(b)}. 
    As new domain prompts are learned, they are added to a domain-specific memory bank.  We then minimize the style similarity between the newly generated domain and previous domains in the memory bank, while maximizing the semantic consistency between generated images and the corresponding category in step \textcolor{red}{(c)}.
    }
    \label{pic:main}
    \vspace{-5.5mm}
\end{figure*}  

\section{Related work}
\subsection{Domain Generalization}
Over the last decade, many efforts have been devoted to domain generalization (DG) to address the OOD issue. 
Currently, DG methods can mainly be categorized into five dimensions, including domain alignment~\cite{ganin2016domain, cheng2024disentangled}, meta-learning~\cite{dou2019domain, wang2020meta, zhang2021adaptive}, data augmentation~\cite{carlucci2019domain, shankar2018generalizing}, disentangled representation learning~\cite{khosla2012undoing, peng2019domain}, and capturing causal relations. 
Despite showing encouraging performances on OOD data, their real-world applications are still limited due to the necessity of having data from other domains.
In this work, we focus on an extreme case in DG: single domain generalization (SDG).
\subsection{Data augmentation for SDG}
SDG focuses on generalizing a model learned from only one source domain to multiple unseen target domains. 
To address this challenging problem, several methods have designed various data augmentation algorithms to enhance the diversity and informativeness of training data.
In particular, some works~\cite{qiao2020learning, volpi2019addressing, volpi2018generalizing, zhao2020maximum} have demonstrated that the method of adversarial domain augmentation (ADA) can be highly effective in enhancing the generalization ability and robustness of the model by synthesizing virtual images during the training process. \cite{zhou2021domain, wang2021learning} operate in the style space defined with style statistics, which can be simply applied to any tasks/models. 
\cite{fan2021adversarially} proposes a generic normalization approach (ASR-Norm) to adapt the data originating from diverse domains when trained with ADA.
Although these methods construct new domains via data augmentation, these methods are not able to cover an arbitrary style that has a large distribution gap with the source domain. 
To address this limitation, we leverage the T2I foundation model to generate a wider variety of domain images, aiming for larger diversity.

\subsection{Text-to-Image model}
T2I model has been widely studied in the context of GANs~\cite{goodfellow2020generative}, with the key developments including models like DALL-E~\cite{ramesh2021zero} and CogView~\cite{ding2021cogview}, which allows users to synthesize novel scenes with unseen compositions and produce vivid images in a myriad of styles.
Recently, remarkable visual results have been achieved by leveraging large-scale auto-regressive~\cite{ramesh2021zero} or diffusion models~\cite{nichol2021glide}. 
\cite{rombach2022high} proposes the latent diffusion model, which applies diffusion model in the latent space of powerful pre-trained autoencoders and uses text with rich semantic information as a condition to guide the generation of images. 
Some work~\cite{clark2023directly, xu2024imagereward} fit a neural network reward function using human scores or relative preferences and then finetune the latent diffusion model.
However, re-training a model with an expanded dataset for each new concept is prohibitively expensive. 
Thus, \cite{gal2022image} proposes a method that does not alter the parameters of the latent diffusion model. Instead, they optimize conditional input prompts.
\section{Methodology}
 
\subsection{Problem Definition} 
Let $\mathcal{D}^\mathcal{S}$ be one labeled source domain and $\mathcal{D}^\mathcal{T}=\{\mathcal{D}_{1}^\mathcal{T},\mathcal{D}_{2}^\mathcal{T},\cdots,\mathcal{D}_{N^\mathcal{T}}^\mathcal{T}\}$ be the multiple target domains, where $N^\mathcal{T}$ is the total number of target domains. 
We denote $\mathcal{D}^\mathcal{S}=\{(\mathbf{x}_i,y_i)\}_{i=1}^{N^s}$, where $\mathbf{x}_i \in \mathcal{X}$ is the $i_{th}$ training example sampled from the input space $\mathcal{X}$, and $y_i \in \mathcal{Y}$ is the corresponding label. In particular, we consider $K$ classes in each domain. 
The $m_{th}$ unseen target domain can also be defined as $\mathcal{D}_{m}^\mathcal{T}=\{(\mathbf{x}_i,y_i)\}_{i=1}^{N^\mathcal{T}_{m}}$, where $N^\mathcal{T}_{m}$ is the number of training samples from the $m_{th}$ domain $\mathcal{D}_{m}^\mathcal{T}$.
Single-domain generalization (SDG) aims to build a model that can perform well on all unseen target domains, through generating images in diverse domain styles to address the domain distribution shift problem.
Note that, this study is verified on solving the image classification problem.

\subsection{Overall Framework of our Method}
This paper proposes a novel Progressive Adversarial Prompt Tuning (PAPT) framework for the pre-trained T2I model, to generate images of a wide range of domain styles, for SDG. 
Our approach aims to learn two sets of abstract prompts instead of the textual description to instruct the diffusion-based T2I foundation model. 
Specifically, the \emph{Category Prompt Tuning} is first adopted to capture categorical features from the source domain data. 
Then, \emph{Adversarial Domain Prompt Tuning} is utilized to learn domain prompts, which model domain styles different from some predefined or previously learned ones in the domain-specific memory bank. 
To generate a sufficiently diverse range of images across different domains automatically, we propose the \emph{Progressive Training Scheme}. 
This mechanism designs a memory bank to store the learned domain prompts, which is iteratively updated by adding the new domain prompt. Specifically, the new domain prompt is progressively and adversarially learned to be different from existing ones.
Finally, by leveraging source domain and the multiple generated domain images together, we can learn a powerful SDG model.

\subsection{Preliminaries} 
We implement our method over Latent Diffusion Models (LDMs)~\cite{rombach2022high}, 
which consist of two core components. Firstly, an auto-encoder is pre-trained on a large dataset. An encoder $\mathbf{f}_E$ learns to map the input images $\mathbf{x}$ into the spatial latent representation $\mathbf{z}_0=\mathbf{f}_E(\mathbf{x})$. The decoder $\mathbf{f}_D$ reconstructs the image from the latent representation, such that $\mathbf{f}_D(\mathbf{f}_E(\mathbf{x}))\xrightarrow{} \mathbf{x}$. The amount of noise added varies according to the timestep $t$ following a variance schedule of $\{\beta_t \in (0,1)\}_{t=1}^{T}$, with $T$ being the total timestep. Then we can add noise $\mathbf{\epsilon}$ to the latent representation $\mathbf{z}_0$ as follows:
\begin{equation}
    \mathbf{z}_{t}=\sqrt{\overline{\alpha}_{t}}\mathbf{z}_{0} + \sqrt{1-\overline{\alpha}_{t}}\mathbf{\epsilon}, \ 
    \mathbf{\epsilon} \in \mathcal{N}\left(0,1\right),
    \label{function:add_noise}
\end{equation} 
where $\mathbf{\epsilon}$ is sampled from Gaussian distribution and $\alpha_t=1-\beta_t$, $\overline{\alpha}_t=\prod_{i=1}^{t}\alpha_i$. 
The second component is a conditional denoising module, which learns a reverse diffusion process that predicts the noise $\hat{\epsilon}$:
\begin{equation}
    \hat{\mathbf{\epsilon}}=\mathbf{\epsilon}_{\theta}(\mathbf{z}_t,t,\Phi(\mathbf{w})),
    \label{funtion:noise_add}
\end{equation}
where $\mathbf{w}$ is the text embedding vector containing the image content descriptions ($i.e.$, the class name), and $\Phi(\cdot)$ is an encoder that maps the condition input of some text description into an intermediate representation $\Phi(\mathbf{w})$. 
Following~\cite{rombach2022high}, it is mapped to the intermediate layers of the UNet~\cite{ronneberger2015u} $\mathbf{\epsilon}_{\theta}(\cdot)$ in Eq.~\textcolor{red}{\ref{funtion:noise_add}} via a cross-attention layer implemented as $\mathrm{Attention}(\mathbf{Q},\mathbf{K},\mathbf{V})=\mathrm{Softmax}(\frac{\mathbf{Q}\mathbf{K}^{\top}}{\sqrt{d}})\cdot \mathbf{V}$, with:
\begin{equation}
    \mathbf{Q}=\mathbf{W}_Q^{(i)} \cdot \varphi_i(\mathbf{z}_t), \mathbf{K}=\mathbf{W}_K^{(i)} \cdot \Phi(\mathbf{w}), \mathbf{V}=\mathbf{W}_V^{(i)} \cdot \Phi(\mathbf{w}).
    \label{function:UNet_crossattention}
\end{equation}
Here, $\mathbf{\varphi}_i(\mathbf{z}_t)$ denotes an intermediate representation of the UNet implementing $\epsilon_{\theta}$, $\mathbf{W}_Q^{(i)},\mathbf{W}_V^{(i)},\mathbf{W}_K^{(i)}$ are projection matrices, and $\Phi(\mathbf{w})$ is the conditional input for the diffusion model. The LDM loss can be written as:
\begin{equation}
    \mathcal{L}_{LDM}=\mathbb{E}_{\mathbf{f}_E\left(\textbf{x}\right),y,\mathbf{\epsilon} \in \mathcal{N}\left(0,1\right),t}\left[\|\mathbf{\epsilon}-\hat{\mathbf{\epsilon}}\|_2^2\right],
    \label{function:ldm_loss}
\end{equation}
where $\|\cdot\|$ is $L2$ norm~\cite{hinton2015distilling} and both $\Phi(\cdot)$ and $\mathbf{\epsilon}_{\theta}$ are jointly optimized via Eq.~\textcolor{red}{\ref{function:ldm_loss}}.

\subsection{Category Prompt Tuning} \label{CPT}
The Category Prompt Tuning aims to capture domain-invariant category information from the source domain data.
In real-world scenarios, some concepts/categories are difficult for the diffusion model to understand through a single class label/name. 
So in the first stage, we use LDMs to train a set of prompts for each category, serving as an abstract representation for the corresponding class. 
Specifically, we denote  $\mathbf{t}^{c}_{k}$ as the input string to facilitate the domain-invariant category description of the $k_{th}$ class as follows, 
\begin{equation}
    \mathbf{t}_k^c = ``a \ photo \ of \ a \ast", \ k \in [1,K].
    \label{function:prompt_generate}
\end{equation}
In Eq.~\textcolor{red}{\ref{function:prompt_generate}}, $K$ is the total number of classes, and `$\ast$' is defined as a `placeholder' indicating the position of class name within the input string. 
Subsequently, the input string $\mathbf{t}_k^c$ follows tokenization and then each token is transformed into the embedding vector. 
Suppose $M_c$ is the length of the tokens corresponding to the input string $\mathbf{t}_k^c$ after the exclusion of the placeholder `$\ast$', then the corresponding tokenized vector can be defined as $\hat{\mathbf{t}}_k^c = [\hat{\mathbf{t}}_{k,1}^c,\cdots, \hat{\mathbf{t}}_{k,j}^c, \cdots, \hat{\mathbf{t}}_{k,M_c}^c$]. Following the embedding function $\varepsilon(\cdot)$ for all the tokens, we then add a learnable category prompt $\mathbf{E}^c_k$ for the $k_{th}$ domain-invariant category description at the position of the placeholder in the embedding space:
\begin{equation}
    \mathbf{w}_k^c=[\varepsilon(\hat{\mathbf{t}}_{k,1}^c),\cdots,\varepsilon(\hat{\mathbf{t}}_{k,j}^c),\cdots,\varepsilon(\hat{\mathbf{t}}_{k,M_c}^c), \mathbf{E}^c_k ].
    \label{function:token_generate}
\end{equation}
Afterward, we use $\Phi(\mathbf{w}_k^c)$ as the category condition input to the cross-attention layer of the UNet $\mathbf{\epsilon}_{\theta}(\cdot)$, directing the prediction of noise while optimizing only $\mathbf{E}^c_k$ with $\mathcal{L}_{LDM}$ in Eq.~\textcolor{red}{\ref{function:ldm_loss}}. 
To be specific, $\Phi(\cdot)$ is the text encoder of CLIP~\cite{radford2021learning} in our experiment. Finally, we can learn $K$ category prompts with each for one class.

\subsection{Adversarial Domain Prompt Tuning}

\paragraph{Image Sampling with Domain Prompt.}
In this paper, we aim to learn a set of abstract domain prompts, where each prompt in the domain-specific memory bank $\mathcal{P}$ is used to instruct the pre-trained diffusion model to generate images in a specific domain style:
\begin{equation}
    \mathcal{P}=\{E^d_1,E^d_2,\cdots,E^d_N\}, \ \mathcal{P} \in \mathbb{R}^{N \times \mathrm{D}},
    \label{function:pooling}
\end{equation}
where $N$ in Eq.~\textcolor{red}{\ref{function:pooling}} represents the number of learned domains in $\mathcal{P}$, and $\mathrm{D}$ represents the prompts dimension.
Given the $i_{th}$ domain prompt $\mathbf{E}_i^d$ in $\mathcal{P}$, we compose a text description to generate an image belonging to $k$-th category as follows:  
\begin{equation}
\begin{array}{cc}
    \mathbf{t}_{k,i}^{c\text{-} d}=``a \ photo \ of \ a \ \ast \ in \ the \ style \ of \ \dagger", 
    \vspace{1.5mm} \\
    where \ k \in [1,K], \ i \in [1, N],
\end{array}
\label{function:prompt_generate_domain}
\end{equation}
where the token embedding of $\ast$ and $\dagger$ will be replaced by our learnable category prompt $\mathbf{E}_k^c$ and domain prompt $\mathbf{E}_i^d$, respectively. 
Similar to Eq.~\textcolor{red}{\ref{function:token_generate}}, $\hat{\mathbf{t}}_{k,i}^{c\text{-} d}$ is defined as a set of tokens corresponding to the input string $\mathbf{t}_{k,i}^{c\text{-} d}$. As a result, the overall token embedding of $\mathbf{w}_{k,i}^{c\text{-} d}$ is formulated as:
\begin{equation}
    \mathbf{w}_{k,i}^{c\text{-} d}=[\varepsilon(\hat{\mathbf{t}}_{k,i,1}^{c\text{-} d}),\cdots,\mathbf{E}^c_k, \cdots,\varepsilon(\hat{\mathbf{t}}_{k,i,M_{c\text{-} d}}^{c\text{-} d}), \mathbf{E}_i^d],
    \label{function:token_generate_all}
\end{equation}
where $M_{c\text{-} d}$ is the length of the tokens in $\hat{\mathbf{t}}_{k,i}^{c\text{-} d}$. 
Finally, we use the pre-trained diffusion model to generate the image belonging to $k_{th}$ category with $i_{th}$ domain style, by feeding the obtained token embedding $\mathbf{w}_{k,i}^{c\text{-} d}$ into the diffusion denoiser in Eq.~\textcolor{red}{\ref{funtion:noise_add}} at every Denoising Diffusion Probabilistic Models (DDPMs)~\cite{ho2020denoising} sampling step. 

\paragraph{Domain Prompt Tuning.} 
Next, we elaborate on how to learn the domain prompt by our proposed Adversarial Domain Prompt Tuning module. 
The goal of this module is to ensure that the domain style of the image generated with the currently learned domain prompt is different from the domain styles stored in the memory bank $\mathcal{P}$. 

\begin{figure}[!htbp]
    \centering
    \includegraphics[width=0.47\textwidth]{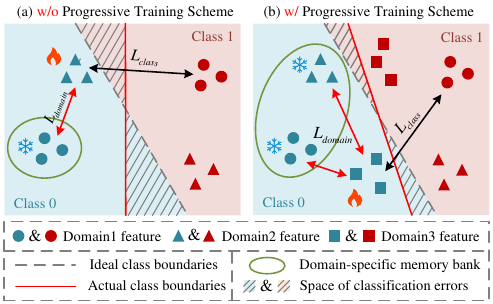}
    \caption{
    This figure shows the impact of the Progressive Training Scheme.
    By comparing the two figures, we found that after using the Progressive Training Scheme, the areas of classification errors have decreased.
    This observation indicates that the more diverse abstract domains we learn, the more likely we are to generate OOD images that are close to the classification boundary.
    }
    \label{pic:method}
    \vspace{-4.5mm}
\end{figure}

First, we set up an initial memory bank of prompts by manually selecting several textual domains and using their corresponding token embeddings as the initial domain prompts in the memory bank. 
In order to learn a new domain prompt $\mathbf{E}_o^d$, we first use the diffusion model to generate an original latent representation $\mathbf{z}_0$ from the $k_{th}$ category by composing the text description following Eq.~\textcolor{red}{\ref{function:prompt_generate_domain}}.
Then, the domain similarity between the generated $\mathbf{z}_0$ and the $i$-th domain style in $\mathcal{P}$ is formulated as the probability that $\mathbf{z}_0$ belongs to the $i_{th}$ domain: 
\begin{equation}  
    \mathbf{p}^d_i(\mathbf{z}_0; \Phi(\mathbf{w}^d_i))=\frac{\exp (<\mathbf{z}_0,\Phi(\mathbf{w}^d_i)>/\tau)}{\sum_{i=1}^{N} \exp (<\mathbf{z}_0, \Phi(\mathbf{w}^d_i)>/\tau)},
    \label{ContrastLearning}
\end{equation}
where we can define $\mathbf{p}^d=[\mathbf{p}^d_1,\cdots,\mathbf{p}^d_i,\cdots,\mathbf{p}^d_N]$ as the collection of probabilities for domain classification. $\tau$ is a hyper-parameter to control the sharpness of the output, and $<\cdot,\cdot>$ is the dot product which can be termed as the cosine similarity as the features are normalized. 

The $\mathbf{w}_i^d$ in Eq.~\textcolor{red}{\ref{ContrastLearning}} represents the token embedding corresponding to the domain-specific textual description $\mathbf{t}_i^d$ in Eq.~\textcolor{red}{\ref{function:domain_specific_string}} of the $i_{th}$ domain style in the $\mathcal{P}$:
\begin{equation}
    \mathbf{t}_i^d = `` \ in \ the \ style \ of \ \dagger", i \in [1,N],
    \label{function:domain_specific_string}
\end{equation}
\begin{equation}
    \mathbf{w}_i^d=[\varepsilon(\hat{\mathbf{t}}_{i,1}^d),\cdots,\varepsilon(\hat{\mathbf{t}}_{i,j}^d),\cdots,\varepsilon(\hat{\mathbf{t}}_{i,M_d}^d), \mathbf{E}^d_i ],
    \label{function:token_generate_domain}
\end{equation}
where $\hat{\mathbf{t}}_i^d$ can be defined as a set of tokens corresponding to the domain-specific textual description $\mathbf{t}_i^d$ and $M_d$ is the length of the tokens in $\hat{\mathbf{t}}_i^d$.
As a result, the adversarial domain loss is formulated to minimize the total style similarity between the $\mathbf{z}_0$ and all the previous domains in the memory bank $\mathcal{P}$. $\mathbf{w}^d_i$ corresponds to each learnable domain prompt $\mathbf{E}_i^d$ in $\mathcal{P}$.
In other words, we want the newly generated domain $\mathbf{E}_o^d$ to be distributionally different from the previous domain in $\mathcal{P}$:
\begin{equation}
    \mathcal{L}_{domain}=-\sum_{i=1}^{N} -\log \mathbf{p}^d_i(\mathbf{z}_0; \Phi(\mathbf{w}^d_i)).
    \label{function:domain_reward}
\end{equation}

To ensure $\mathbf{z}_0$ still maintains the category semantics after optimizing with the adversarial domain loss, we further propose a category consistency loss, which maximizes the semantic consistency between $\mathbf{z}_0$ and the categorical embedding $\mathbf{w}_k^c$ in Eq.~\textcolor{red}{\ref{function:token_generate}}: 
\begin{equation}
    \mathcal{L}_{class}=-\log \mathbf{p}_k^c(\mathbf{z}_0; \Phi(\mathbf{w}^c_k))),
    \label{funtion:class_reward}
\end{equation}
where the $\mathbf{p}^c_k$ represent the probability of $\mathbf{z}_0$ belonging to the $k_{th}$ category. We calculate the $\mathbf{p}^c_k$ with the same as Eq.~\textcolor{red}{\ref{ContrastLearning}} by $\mathbf{z}_0$ and $\Phi(\mathbf{w}^c_k)$.
The final adversarial domain prompt tuning loss is written as:
\begin{equation}
    \mathcal{L}_{all}=\mathcal{L}_{class}+\lambda\mathcal{L}_{domain},
    \label{function:reward_all}
\end{equation}  
where $\lambda$ in Eq.~\textcolor{red}{\ref{function:reward_all}} is a hyper-parameter. It balances the trade-off between training the new domain prompt distributionally different from existing ones and preserving its key categorical semantic information.  
Optimizing Eq.~\textcolor{red}{\ref{function:reward_all}} requires sampling $\mathbf{z}_0$ and back-propagating the gradient through the entire sequence of sampling steps, which is impractical. 
As a result, we propose using a one-step approximation of $\mathbf{z}_0$. Specifically, we first sample an inter-mediate state $\mathbf{z}_t$ with a relatively low noise level, and then conduct a one-step approximation of $\mathbf{z}_0$ with $\mathbf{z}_t$, and the loss gradient is only back-propagated from the $t$-th to the $0$-th step:
\begin{equation}
    \mathbf{z}_{0}=\frac{1}{\sqrt{\overline{\alpha}_{t}}}\left(\mathbf{z}_{t}-\sqrt{1-\overline{\alpha}_{t}}\mathbf{\epsilon}_{\theta}\left(\mathbf{z}_{t},t,\Phi\left(\mathbf{w}_{k,i}^{c\text{-} d}\right)\right)\right).
    \label{function:denoising_second}
\end{equation}

\subsection{Progressive Training Scheme}
To automatically generate images with diverse domain styles, we need to progressively update $\mathcal{P}$ by continuously adding new domain prompts. 
To achieve this, we propose a Progressive Training Scheme: after a certain number of prompt tuning iterations, the currently learned domain prompt $\mathbf{E}_o^d$ is added to $\mathcal{P}$, allowing $\mathcal{P}$ to be gradually updated with new styles as follows:
\begin{equation}
    \mathcal{P} \leftarrow \mathcal{P} \cup \{\mathbf{E}^d_o\}.
\end{equation}
Fig.~\textcolor{red}{\ref{pic:method}} provides a sample illustration. 
The purpose of this training scheme is to ensure that each time when we train the new domain $\mathbf{E}_{o}^d$ using adversarial domain loss in Eq.~\textcolor{red}{\ref{function:domain_reward}}, we can also increase the distance between the new domain $\mathbf{E}_{o}^d$ and the previously added domain $\mathbf{E}_i^d$ in $\mathcal{P}$. 
In this way, each training domain will be distributionally different from previous ones. 
Thus, we have a higher probability of learning more challenging abstract domains.

\section{Experiments}
\subsection{Datasets and Evaluation Protocols}
Following previous works~\cite{qu2023modality, shinunknown},
we adopt five commonly used benchmark datasets in DG tasks for evaluation:
PACS~\cite{li2017deeper} (4 domains, 9,991 samples, 7 classes), VLCS~\cite{li2017deeper} (4 domains, 10,729 samples, 5 classes), OfficeHome~\cite{venkateswara2017deep} (4 domains, 15,588 samples, 65 classes), DomainNet~\cite{peng2019moment} (6 domains, 586,575 samples, 345 classes) and TerraIncognita~\cite{beery2018recognition} (4 domains, 24,788 samples, 10 classes). 
To ensure reliable results, we calculate the average performance across multiple experiments (Avg-acc).

\subsection{Implementation details}
In our implementation, we adopt the ResNet-18~\cite{he2016deep} for PACS and VLCS and ResNet-50~\cite{he2016deep} for OfficeHome pre-trained on ImageNet~\cite{deng2009imagenet} as backbone for the SDG setting, follows~\cite{qu2023modality, shinunknown} for a fair comparison.
The models are trained for 50 epochs using a batch size of 64 and a learning rate of 0.001 for each stage. We utilize the SGD optimizer and conduct the training on a single NVIDIA RTX3090 GPU. 
In our proposed method, the parameter $N$ in Eq.~\ref{function:pooling} is set to $5$. The parameter $\lambda$ in Eq.~\ref{function:reward_all} is set to $10$. 
To ensure the reliability of our method, we independently repeat all experiments five times and report the average results.

\begin{table}[!htbp]
\vspace{-2.5mm}
\caption{
SDG results (\%) on PACS.
One domain is used as the source domain and the others are used as the target domain.
}
\renewcommand\arraystretch{0.9}
\scalebox{0.845}{
\centering
\footnotesize
\begin{tabular}{l|c|cccc|>{\columncolor{gray!30}}c}
\hline
\textbf{Method} &\textbf{Venue} &\textbf{P} &\textbf{A} &\textbf{C} &\textbf{S} &\textbf{Avg.}  \\
\hline
Augmix~\cite{hendrycks2019augmix}      
& ICLR'20      &38.30             &66.54 &70.16             &52.48             &56.87 \\
RSC~\cite{huang2020self}             
& ECCV'20      &41.60             &73.40 &75.90             &56.20             &61.80 \\
L2D~\cite{wang2021learning}
& ICCV'21      &52.29             &76.91 &77.88             &53.66             &65.18 \\
RSC+ASR~\cite{huang2020self}      
& CVPR'21      &54.60             &76.70 &79.30             &61.60             &68.10 \\
pAdaIn~\cite{nuriel2021permuted}      
& CVPR'21      &33.66             &64.96 &65.24             &32.04             &49.98 \\
ERM~\cite{vapnik2013nature}         
& ICLR'21      &33.65             &65.38 &64.20             &34.15             &49.34 \\
Mixstyle~\cite{zhou2021domain}    
& ICLR'21      &37.44             &67.60 &70.38             &34.57             &52.50 \\
EFDMix~\cite{zhang2022exact}      
& CVPR'22      &42.50             &63.20 &73.90             &38.10             &54.40 \\
DSU~\cite{li2022uncertainty}         
& ICLR'22      &42.10             &71.54 &74.51             &47.75             &58.97 \\
ACVC~\cite{cugu2022attention}
& CVPRW'22     &48.05             &73.68 &77.39             &55.30             &63.61 \\
MAD~\cite{qu2023modality}
& CVPR'23      &52.95             &75.51 &77.25             &57.75             &65.87 \\
P-RC~\cite{choi2023progressive}
& CVPR'23      &57.11             &76.98 &78.54             &62.89             &68.88 \\
Meta-Casual~\cite{chen2023meta}
& CVPR'23      &59.60             &77.13 &\underline{80.14} &62.55             &69.86 \\
ITTA~\cite{chen2023improved}  
& CVPR'23      &56.50             &\underline{78.40} &79.80             &60.70             &68.80 \\
Prompt-Driven~\cite{li2024prompt} 
& CVPR'24      &\underline{60.09} &\textbf{78.77}    &\textbf{82.69}    &\underline{62.94}             &\underline{71.12} \\
\hline
\textbf{PAPT (Ours)}         
& -            &\textbf{62.12}    &69.84 &78.57             &\textbf{80.34}    &\textbf{72.72} \\ 
\hline
\end{tabular}}
\label{tab:pacs}
\vspace{-3.5mm}
\end{table}
\begin{table}[!htbp]
\vspace{-2.5mm}
\caption{
SDG results (\%) on VLCS. 
One domain is used as the source domain and the others are used as the target domain.
}
\centering
\renewcommand\arraystretch{0.9}
\scalebox{0.905}{
\footnotesize
\begin{tabular}{l|c|cccc|>{\columncolor{gray!30}}c}
\hline
\textbf{Method} &\textbf{Venue} &\textbf{V} &\textbf{L} &\textbf{C} &\textbf{S} &\textbf{Avg.}  \\
\hline
Augmix~\cite{hendrycks2019augmix}      
& ICLR'20     &75.25 &59.52 &45.90 &57.43 &59.53 \\
ERM~\cite{vapnik2013nature}
& ICLR'21     &\textbf{76.72} &58.86 &44.95 &57.71 &59.56 \\
pAdaIn~\cite{nuriel2021permuted}
& CVPR'21     &76.03 &65.21 &43.17 &57.94 &60.59 \\
Mixstyle~\cite{zhou2021domain}
& ICLR'21     &75.73 &61.29 &44.66 &56.57 &59.56 \\
EFDMix~\cite{zhang2022exact}
& CVPR'22     &72.35 &61.41 &\underline{52.34} &\underline{63.28} &62.33 \\
DSU~\cite{li2022uncertainty}
& ICLR'22     &76.93 &69.20 &46.54 &58.36 &62.76 \\
ACVC~\cite{cugu2022attention}
& CVPRW'22    &76.15 &61.23 &47.43 &60.18 &61.25 \\
MAD~\cite{qu2023modality}
& CVPR'23     &\underline{76.15} &\underline{69.36} &48.04 &61.74 &\underline{63.82} \\
\hline
\textbf{PAPT (Ours)}          & -           &73.16 &\textbf{74.69} &\textbf{69.66} &\textbf{75.96} &\textbf{73.37} \\ 
\hline
\end{tabular}}
\label{tab:vlcs}
\vspace{-3.5mm}
\end{table}
\begin{table}[!htbp]
\small
\begin{center}
\caption{
SDG results (\%) on OfficeHome. 
One domain is used as the source domain and the others are used as the target domain.
}
\label{tab:multi-source-officehome}
\renewcommand\arraystretch{0.9}
\scalebox{0.80}{
\begin{tabular}{l|c|cccc|>{\columncolor{gray!30}}c}
\hline
\textbf{Method}  &\textbf{Venue}    &\textbf{A}     &\textbf{C}     &\textbf{P}     &\textbf{R}     &\textbf{Avg.}  \\ 
\hline
DANN~\cite{ganin2016domain}       
& IJCAI'16    & 55.20 & 49.30  & 48.40  & 58.40 & 52.80 \\
CORAL~\cite{sun2016deep}       
& ICCV'16     & 55.60 & 52.80  & 50.30  & 59.40 & 54.50 \\
IRM~\cite{arjovsky2019invariant}
& -           & 54.90 & 53.20  & 48.60  & 59.20 & 54.00 \\
MMD~\cite{li2018domain}
& ICCV'18     & 55.10 & 52.00  & 50.30  & 59.30 & 54.20 \\
OrgMixup~\cite{zhang2018mixup}
& ICLR'18     & 56.00 & 54.40  & 50.40  & 61.00 & 55.50 \\
Mixup~\cite{yan2020improve}
& -           & 55.50 & 54.10  & 49.40  & 59.40 & 54.60 \\
CutMix~\cite{yun2019cutmix}
& CVPR'19     & 53.50 & 52.20  & 47.70  & 60.20 & 53.40 \\
CDANN~\cite{ganin2016domain}
& -           & 55.20 & 49.90  & 47.60  & 58.60 & 52.80 \\
GroupDRO~\cite{sagawa2019distributionally}    
& ICLR'20     & 55.10 & 52.00  & 50.30  & 59.30 & 54.20 \\
MTL~\cite{blanchard2021domain}
& JMLR'21     & 55.30 & 53.30  & 49.00  & 60.40 & 54.50 \\
ARM~\cite{zhang2021adaptive}
& NeurIPS'21  & 55.00 & 51.60  & 47.30  & 59.30 & 53.30 \\
VREx~\cite{krueger2021out}        
& ICML'21     & 55.50 & 52.60  & 49.10  & 59.30 & 54.10 \\
Mixstyle~\cite{zhou2021domain}
& ICLR'21     & 44.30 & 29.80  & 33.60  & 48.50 & 39.00 \\
ERM~\cite{vapnik2013nature}
& ICLR'21     & 55.60 & 52.80  & 50.30  & 59.40 & 54.50 \\
SAM~\cite{li2018domain}
& ICLR'21     & 56.90 & 53.80  & 50.90  & 61.50 & 55.80 \\
SagNet~\cite{nam2021reducing}
& CVPR'21     & 56.90 & 53.40  & 50.80  & 61.20 & 55.60 \\
Fishr~\cite{rame2022fishr}
& ICML'22     & 55.10 & 51.20  & 49.20  & 59.90 & 53.90 \\
RIDG~\cite{chen2023domain}
& ICCV'23     & 56.80 & 55.40  & 50.50  & 60.90 & 55.90 \\
SAGM~\cite{wang2023sharpness}
& CVPR'23     & 57.70 & 54.80  & 51.50  & 61.40 & 56.30 \\
ITTA~\cite{chen2023improved}
& CVPR'23     & 56.00 & 51.50  & 50.50  & \underline{61.60} & 54.90 \\
UDIM~\cite{shinunknown}
&ICLR'24      & \underline{58.50} & \underline{55.70}  & \underline{54.50}  & \textbf{64.50} & \underline{58.30} \\ 
\hline              
\textbf{PAPT (Ours)}  
& -           & \textbf{59.81} & \textbf{68.30} &\textbf{59.12} & 60.87 &\textbf{62.03}   \\ 
\hline\end{tabular}}
\end{center}
\vspace{-6.5mm}
\end{table}

\subsection{Results on Single Domain Generalization}
We compare our method with baseline method~\cite{vapnik2013nature}, the data augmentation based method~\cite{nuriel2021permuted, zhou2021domain, zhang2022exact, hendrycks2019augmix, li2022uncertainty, cugu2022attention, chen2023domain, qu2023modality}, the domain invariant representation based method~\cite{ganin2016domain, li2018domain, arjovsky2019invariant, krueger2021out, blanchard2021domain, sun2016deep, huang2020self}, the feature disentanglement based method~\cite{nam2021reducing}, the distributionally robust optimization based method~\cite{sagawa2019distributionally}, the gradient operation based method~\cite{huang2020self, rame2022fishr} and the flatness-aware based methods~\cite{li2018domain, zhuang2022surrogate, wang2023sharpness, shinunknown}.

\begin{table*}[!htbp]
\small
\begin{center}
\caption{
Multi-source DG results (\%) on PACS, VLCS, OfficeHome, TerraInc, and DomainNet benchmark datasets. 
}
\vspace{-2.0mm}
\label{tab:multi-DomainBed}
\renewcommand\arraystretch{1.0}
\scalebox{0.965}{
\begin{tabular}{l@{\hspace{16pt}}|@{\hspace{16pt}}c@{\hspace{16pt}}|@{\hspace{16pt}}c@{\hspace{16pt}}c@{\hspace{16pt}}c@{\hspace{16pt}}c@{\hspace{16pt}}c@{\hspace{16pt}}|>{\hspace{13pt}}>{\columncolor{gray!30}}c<{\hspace{13pt}}}
\hline
\textbf{Method} &\textbf{Venue} &\textbf{PACS} &\textbf{VLCS} &\textbf{OfficeHome} &\textbf{TerraInc}    &\textbf{DomainNet}  &\textbf{Avg.}  \\ 
\hline
DANN~\cite{ganin2016domain}       
& IJCAI'16    
& 83.60\tsb{\(\pm\)0.40} & 78.60\tsb{\(\pm\)0.40} & 65.90\tsb{\(\pm\)0.60} 
& 46.70\tsb{\(\pm\)0.50} & 38.30\tsb{\(\pm\)0.40} & 62.60 \\
CORAL~\cite{sun2016deep}       
& ICCV'16     
& 86.20\tsb{\(\pm\)0.30} & 78.80\tsb{\(\pm\)0.60} & 68.70\tsb{\(\pm\)0.30} 
& 47.60\tsb{\(\pm\)1.00} & 41.50\tsb{\(\pm\)0.10} & 64.50 \\
MLDG~\cite{li2018learning}        
& AAAI'18     
& 84.90\tsb{\(\pm\)1.00} & 77.20\tsb{\(\pm\)0.40} & 66.80\tsb{\(\pm\)0.60} 
& 47.70\tsb{\(\pm\)0.90} & 41.20\tsb{\(\pm\)0.10} & 63.60 \\
GroupDRO~\cite{sagawa2019distributionally}    
& ICLR'20     
& 84.40\tsb{\(\pm\)0.80} & 76.70\tsb{\(\pm\)0.60} & 66.00\tsb{\(\pm\)0.70} 
& 43.20\tsb{\(\pm\)1.10} & 33.30\tsb{\(\pm\)0.20} & 60.70 \\
RSC~\cite{huang2020self}         
& ECCV'20     
& 85.20\tsb{\(\pm\)0.90} & 77.10\tsb{\(\pm\)0.50} & 65.50\tsb{\(\pm\)0.90} 
& 46.60\tsb{\(\pm\)1.00} & 38.90\tsb{\(\pm\)0.50} & 62.70 \\
VREx~\cite{krueger2021out}        
& ICML'21     
& 84.90\tsb{\(\pm\)0.60} & 78.30\tsb{\(\pm\)0.20} & 66.40\tsb{\(\pm\)0.60}
& 46.40\tsb{\(\pm\)0.60} & 33.60\tsb{\(\pm\)2.20} & 61.90 \\
Mixstyle~\cite{zhou2021domain}
& ICLR'21     
& 85.20\tsb{\(\pm\)0.30} & 77.90\tsb{\(\pm\)0.50} & 60.40\tsb{\(\pm\)0.20} 
& 44.00\tsb{\(\pm\)0.70} & 34.00\tsb{\(\pm\)0.10} & 60.30 \\
ERM~\cite{vapnik2013nature}
& ICLR'21     
& 85.50\tsb{\(\pm\)0.20} & 77.30\tsb{\(\pm\)0.40} & 66.50\tsb{\(\pm\)0.30} 
& 46.10\tsb{\(\pm\)1.80} & 43.80\tsb{\(\pm\)0.10} & 63.90 \\
SAM~\cite{li2018domain}
& ICLR'21     
& 85.80\tsb{\(\pm\)0.20} & 79.40\tsb{\(\pm\)0.10} & 69.60\tsb{\(\pm\)0.10} 
& 43.30\tsb{\(\pm\)0.70} & 44.30\tsb{\(\pm\)0.00} & 64.50 \\
SagNet~\cite{nam2021reducing}
& CVPR'21     
& 86.30\tsb{\(\pm\)0.20} & 77.80\tsb{\(\pm\)0.50} & 68.10\tsb{\(\pm\)0.10} 
& 48.60\tsb{\(\pm\)1.00} & 40.30\tsb{\(\pm\)0.10} & 64.20 \\
Miro~\cite{cha2022domain}
& ECCV'22     
& 85.40\tsb{\(\pm\)0.40} & 79.00\tsb{\(\pm\)0.00} & 70.50\tsb{\(\pm\)0.40} 
& 50.40\tsb{\(\pm\)1.10} & 44.30\tsb{\(\pm\)0.20} & 65.90 \\
GSAM~\cite{zhuang2022surrogate}
& ICLR'22     
& 85.90\tsb{\(\pm\)0.10} & 79.10\tsb{\(\pm\)0.20} & 69.30\tsb{\(\pm\)0.00} 
& 47.00\tsb{\(\pm\)0.80} & 44.60\tsb{\(\pm\)0.20} & 65.10 \\
SAGM~\cite{wang2023sharpness}
& CVPR'23     
& 86.60\tsb{\(\pm\)0.20} & \underline{80.00\tsb{\(\pm\)0.30}} & 70.10\tsb{\(\pm\)0.20} 
& 48.80\tsb{\(\pm\)0.90} & 45.00\tsb{\(\pm\)0.20} & 66.10 \\
DomainDrop~\cite{guo2023domaindrop}
& ICCV'23     
& 87.90\tsb{\(\pm\)0.30} & 79.80\tsb{\(\pm\)0.30} & 68.70\tsb{\(\pm\)0.10} 
& 51.50\tsb{\(\pm\)0.40} & 44.40\tsb{\(\pm\)0.50} & 66.50 \\
GMDG~\cite{tan2024rethinking}
& CVPR'24     
& 85.60\tsb{\(\pm\)0.30} & 79.20\tsb{\(\pm\)0.30} & 70.70\tsb{\(\pm\)0.20} 
& 51.10\tsb{\(\pm\)0.90} & 44.60\tsb{\(\pm\)0.10} & 66.30 \\
SMOS~\cite{luo2024grounding}
& CVPR'24     
& 89.40\tsb{\(\pm\)0.30} & 79.80\tsb{\(\pm\)0.10} & \underline{71.60\tsb{\(\pm\)0.10}} 
& \textbf{55.40\tsb{\(\pm\)0.40}} & 45.30\tsb{\(\pm\)0.00} & \underline{68.30} \\
RES~\cite{huangrepresentation}
& ECCV'24     
& \underline{90.00\tsb{\(\pm\)0.30}} & 79.80\tsb{\(\pm\)0.20} & \textbf{71.80\tsb{\(\pm\)0.30}} 
& 51.40\tsb{\(\pm\)0.60} & \underline{46.70\tsb{\(\pm\)0.20}} & 67.90 \\
\hline              
\textbf{PAPT (Ours)}    
& - 
& \textbf{91.69\tsb{\(\pm\)0.44}} & \textbf{80.84\tsb{\(\pm\)0.32}} & 71.21\tsb{\(\pm\)0.53} 
& \underline{54.27\tsb{\(\pm\)1.01}} & \textbf{50.18\tsb{\(\pm\)0.47}} & \textbf{69.64} \\ 
\hline
\end{tabular}}
\end{center}
\vspace{-8.0mm}
\end{table*}

Tab.~\ref{tab:pacs}, Tab.~\ref{tab:vlcs}, and Tab.~\ref{tab:multi-source-officehome} show the comparison results on PACS, VLCS, and OfficeHome, respectively. 
Notably, our proposed method outperforms the state-of-the-art method by 1.60\%, 9.55\%, and 3.73\% on Avg-acc on all benchmark datasets. 
Especially when taking "Sketch" (PACS), "Caltech101" (VLCS), "SUN09" (VLCS), and "Clipart" (OfficeHome) as the source domain, our PAPT leads to 17.40\%, 21.62\%, 14.22\% and 12.60\% improvement in Avg-acc.
It is important to highlight that the domain shift in the VLCS dataset mainly comes from background and viewpoint changes.
In other words, the domain label does not convey any semantic information.
Since the result on VLCS (63.82\%$\rightarrow$73.37\%) further proves our proposed method does not rely on the domain labels from the source domain, it can be applied to other robustness tasks that do not contain domain labels (VLCS and TerraInc). 
To sum up, our proposed method has two main advantages: 
(1) We fully leverage the flexibility of learnable domain prompts to represent abstract and challenge domains instead of relying on rigid textual domains.
(2) Unlike existing data augmentation methods, We propose a novel progressive training scheme to harness the potential of the T2I model to generate diverse images automatically.
Further mitigating the domain discrepancies between source and target domains.

\subsection{Results on Multiple Domain Generalization}
We extend our proposed method to multi-source DG setting by extracting the same images as input from various source domains to train class prompt $\mathbf{E}_k^c$. Then keep the subsequent training process consistent with the SDG setting.
We compare our method with several popular DG methods.

Tab.~\ref{tab:multi-DomainBed} shows the multi-source DG results on all the DG benchmark datasets with ResNet-50~\cite{he2016deep} as the backbone.
Notably, our proposed method outperforms the state-of-the-art method SMOS by 1.34\% on Avg-acc.
(1) This further demonstrates that our proposed method not only benefits single-domain generalization but also enhances multi-source domain generalization; 
(2) In the challenging case with large style discrepancies, such as DomainNet, our PAPT can even gain significant performance than SMOS by 4.88\%.
(3) The representative data augmentation method, $e.g.$ Mixstyle, shows the limited performance with ERM on VLCS (77.30\%$\rightarrow$77.90\%) and TerraInc (43.80\%$\rightarrow$34.00\%) dataset. 
We speculate that this may result from the distinct types of domain shifts, which typically manifest as differences in viewpoint and background in VLCS and TerraInc, rather than as variations in texture and style commonly observed in other DG datasets.
Notably, our proposed PAPT substantially improves the ERM of $3.54\%$ and $6.38\%$ on two datasets, respectively.
This observation indicates that compared to the existing data augmentation based methods, 
our PAPT not only alleviates the negative impacts of style shifts but also guarantees a provable generalization performance under varying viewpoint conditions.

\begin{figure*}[!htbp]
    \centering
    \includegraphics[width=1.0 \textwidth]{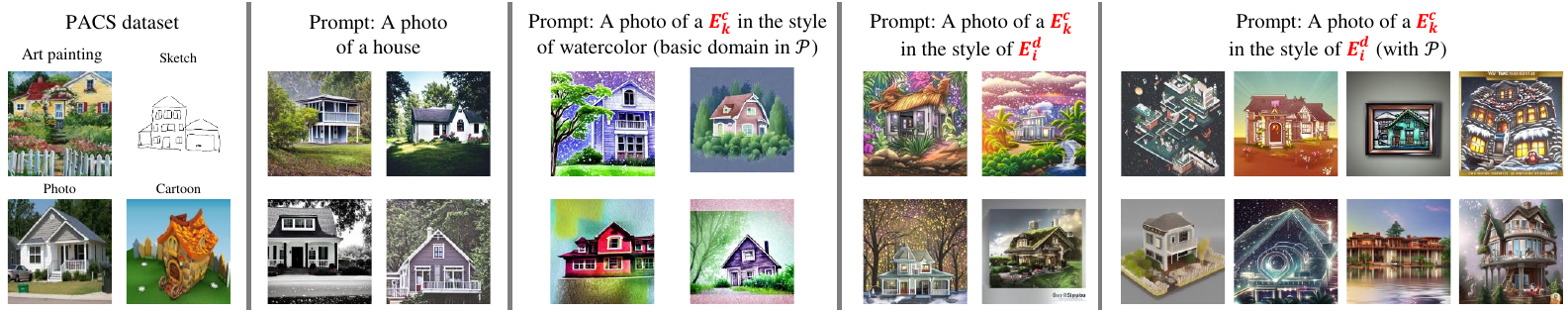}
    \vspace{-7.0mm}
    \caption{
    This figure shows the training data generated by the T2I model under different input conditions, corresponding to each row in the Tab.~\textcolor{red}{\ref{tab:ablation_study}} and Tab.~\textcolor{red}{\ref{tab:ablation_study_vlcs}}.
    It can be observed that our proposed method learns some abstract domains that are difficult to describe with language after replacing the manually designed textual domains with the domain prompt $\mathbf{E}_i^d$. 
    Furthermore, after applying the Progressive Training Scheme, the T2I model generates a sufficiently diverse range of images across different domains.
    }
    \label{pic:experiment_ablution}
    \vspace{-4.5mm}
\end{figure*} 

\subsection{Ablation Study}
To validate the effectiveness of each component of our proposed PAPT, we conduct ablation studies in Tab.~\textcolor{red}{\ref{tab:ablation_study}} and Tab.~\textcolor{red}{\ref{tab:ablation_study_vlcs}}.
Our baseline (see $1^{st}$ row) only uses the source domain as training data. 
In $2^{nd}$ row, we use a simple description ($``a \ photo \ of \ a \ [class]"$) as the input of \textbf{T2I} model to generate additional training images.
In $3^{rd}$ row, we train category prompt $\mathbf{E}_k^c$ using the Category Prompt Tuning \textbf{(CPT)}, and generate images with some textual domains. 
In $4^{th}$ row, we train a domain prompt $\mathbf{E}_i^d$ using the Adversarial Domain Prompt Tuning \textbf{(AD)} and generate images with only one $\mathbf{E}_i^d$ and textual domains.
In $5^{th}$ row, we incorporate the Progressive Training Scheme \textbf{(PTS)} to generate images with diverse domain prompts in $\mathcal{P}$.

\begin{table}[!htbp]
\caption{Ablation studies on individual components of our proposed method on PACS dataset with SDG setting.}
\vspace{-2.0mm}
\centering
\begin{adjustbox}{max width=\textwidth}
\footnotesize
\renewcommand\arraystretch{0.9}
\scalebox{0.90}{
\begin{tabular}{cccc|cccc|>{\columncolor{gray!30}}c}
\hline
\textbf{T2I}  &\textbf{CPT}  &\textbf{AD} &\textbf{PTS} &\textbf{P}  &\textbf{A}  &\textbf{C}  &\textbf{S}  &\textbf{Avg.}         \\
\hline
\ding{55}   &\ding{55}   &\ding{55}  &\ding{55} &34.00  &58.60  &66.40  &27.50  &46.60        \\
\ding{51}   &\ding{55}   &\ding{55}  &\ding{55} &40.23  &59.17  &68.80  &66.91  &58.78        \\
\ding{51}   &\ding{51}   &\ding{55}  &\ding{55} &55.55  &68.87  &71.56  &69.36  &66.34        \\
\ding{51}   &\ding{51}   &\ding{51}  &\ding{55} &58.72  &69.00  &75.61  &77.55  &70.22        \\
\ding{51}   &\ding{51}   &\ding{51}  &\ding{51} &\textbf{62.12}  &\textbf{69.84}  &\textbf{78.57}  &
\textbf{80.34}  &\textbf{72.72}        \\
\hline
\end{tabular}
}
\end{adjustbox}

\label{tab:ablation_study}
\vspace{-4.0mm}
\end{table}
\begin{table}[!htbp]
\caption{Ablation studies on individual components of our proposed method on VLCS dataset with SDG setting.}
\vspace{-2.0mm}
\centering
\begin{adjustbox}{max width=\textwidth}
\footnotesize
\renewcommand\arraystretch{0.9}
\scalebox{0.90}{
\begin{tabular}{cccc|cccc|>{\columncolor{gray!30}}c}
\hline
\textbf{T2I}  &\textbf{CPT} &\textbf{AD} &\textbf{PTS} &\textbf{V}  &\textbf{L}  &\textbf{C}   &\textbf{S}   &\textbf{Avg.}   \\
\hline
\ding{55}   &\ding{55}  &\ding{55}  &\ding{55} &71.81       &61.06       &52.60        &62.32        &61.95           \\
\ding{51}   &\ding{55}  &\ding{55}  &\ding{55} &72.08       &69.31       &60.75        &70.95        &68.28           \\
\ding{51}   &\ding{51}  &\ding{55}  &\ding{55} &72.68       &70.44       &61.54        &71.70        &69.09           \\
\ding{51}   &\ding{51}  &\ding{51}  &\ding{55} &72.94       &73.95       &67.94        &75.20        &72.51           \\
\ding{51}   &\ding{51}  &\ding{51}  &\ding{51} &\textbf{73.16}       &\textbf{74.69}       &\textbf{69.66}        &\textbf{75.96}        &\textbf{73.37}           \\
\hline
\end{tabular}
}
\end{adjustbox}
\label{tab:ablation_study_vlcs}
\vspace{-4.0mm}
\end{table}
\begin{table}[!htbp]
\caption{
The impact of manually designed textual domain initialization in $\mathcal{P}$ on PACS, VLCS, and OfficeHome with SDG setting.
}
\vspace{-2.0mm}
\centering
\begin{adjustbox}{max width=\textwidth}
\footnotesize
\scalebox{1.14}{
\begin{tabular}{c|@{\hspace{10pt}}c@{\hspace{10pt}}c@{\hspace{10pt}}c@{\hspace{10pt}}|>{\columncolor{gray!30}}c}
\hline
\textbf{Idx} &\textbf{PACS}  &\textbf{VLCS} &\textbf{OfficeHome}  &\textbf{Avg.}   \\
\hline
\multicolumn{5}{l}{Initialization with watercolor, oil painting and pixlate.}      \\
\hline
1       &72.72\tsb{\(\pm\)0.32}  &73.37\tsb{\(\pm\)0.21} & 62.03\tsb{\(\pm\)0.38}  &73.05        \\
\hline
\multicolumn{5}{l}{Initialization with ink drawing, cyber art and realism.}        \\
\hline
2       &72.13\tsb{\(\pm\)0.19}  &73.05\tsb{\(\pm\)0.28} & 61.99\tsb{\(\pm\)0.44}  &72.59        \\
\hline
\multicolumn{5}{l}{initialization with printmaking, line art and pastel.}          \\
\hline
3       &72.57\tsb{\(\pm\)0.24}  &73.12\tsb{\(\pm\)0.15} & 62.08\tsb{\(\pm\)0.36}  &72.85        \\
\hline
\end{tabular}
}
\end{adjustbox}
\label{tab:initialization_analyse}
\vspace{-3.0mm}
\end{table}

\begin{figure}[!htbp]
    \centering
    \includegraphics[width=0.48\textwidth]{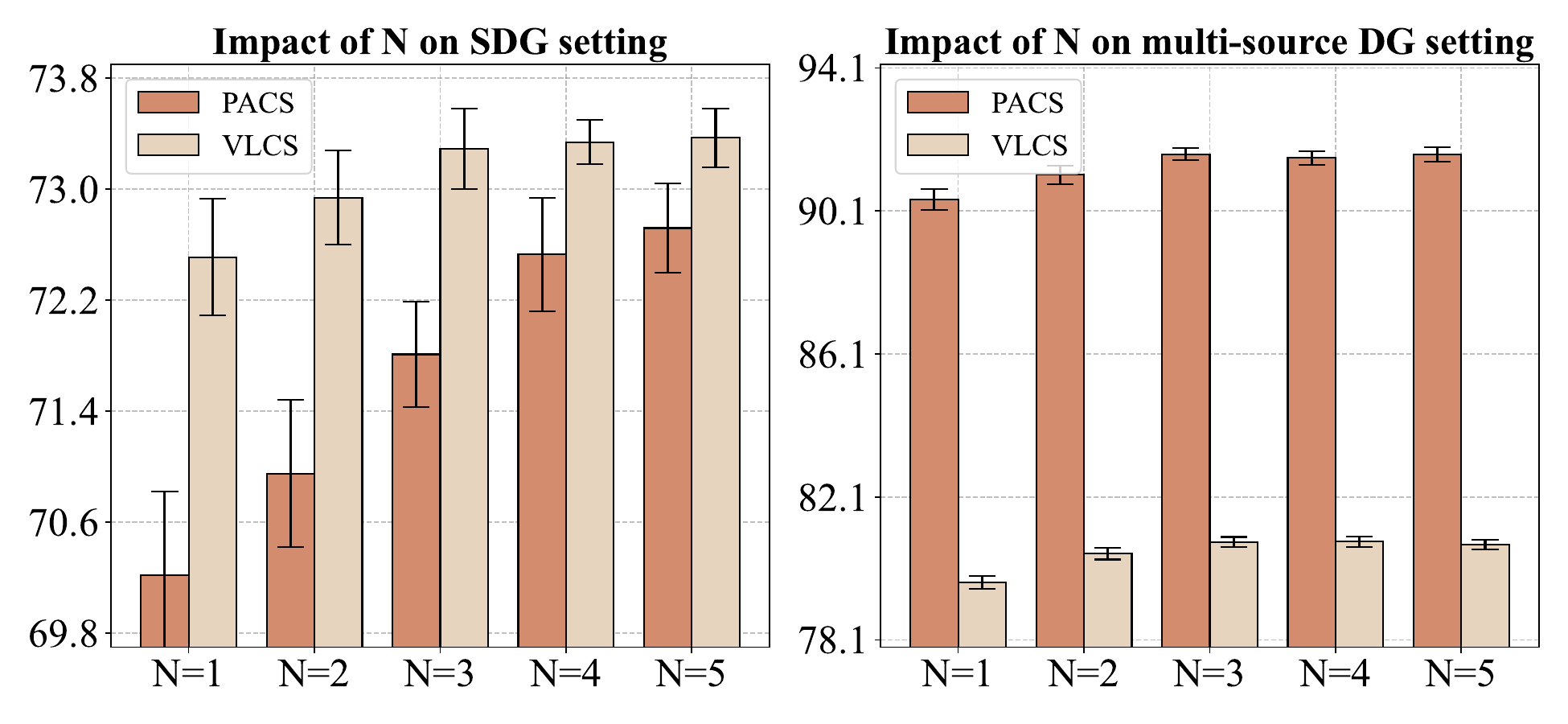}
    \vspace{-6.5mm}
    \caption{
    Impact of domain prompt number in Eq.~\ref{function:pooling} $N$ on PACS and VLCS dataset on SDG setting and multi-source DG setting. 
    }
    \label{pic:parameter_analysis_N}
    \vspace{-5.5mm}
\end{figure} 
\noindent \textbf{The Effectiveness of Adversarial Domain Prompt Tuning (AD).} 
Our Adversarial Domain Prompt Tuning substantially improves the model performance of 0.13\%$\sim$8.19\% across each domain and 3.88\% on Avg-acc (see $3^{rd}$ and $4^{th}$ row in Tab.~\textcolor{red}{\ref{tab:ablation_study}}) on PACS dataset.
In contrast, the manually designed textual domain demonstrates limited performance as some domain characteristics are too abstract to describe with words. 
Unlike relying on rigid textual input, our AD fully leverages the flexibility of learnable domain prompts to more effectively characterize the diverse abstract styles, resulting in a generalization performance gain.
Meanwhile, existing data augmentation methods that mainly introduce diverse styles obtain relatively slight performance gains on VLCS dataset compared to PACS ones. 
However, our AD significantly improves the model performance of 0.28\%$\sim$6.40\% across each domain and 3.42\% on Avg-acc (see $3^{rd}$ and $4^{th}$ row in Tab.~\textcolor{red}{\ref{tab:ablation_study_vlcs}}).
This observation indicates our AD can better model the changes in background and viewpoint and mitigate the distributional difference under these conditions with the powerful pre-trained T2I model.
The results basically indicate the advantage of AD across different domain shifts.
Fig.~\ref{pic:experiment_ablution} also shows some images generated by our proposed method.

\noindent \textbf{The Effectiveness of Progressive Training Scheme (PTS).} 
Our Progressive Training Scheme achieves improvements of 0.84\%$\sim$3.40\% across each domain and 2.50\% on Avg-acc (see $4^{th}$ row and $5^{th}$ row in Tab.~\textcolor{red}{\ref{tab:ablation_study}}) on PACS dataset and 0.22\%$\sim$1.72\% across each domain and 0.83\% on Avg-acc (see $4^{th}$ row and $5^{th}$ row in Tab.~\textcolor{red}{\ref{tab:ablation_study_vlcs}}) on VLCS dataset. 
This observation indicates that by ensuring uniformly well performance across sufficiently diverse domains, our PAPT can further shrink the distribution discrepancies between source and target domains.
To visualize the effectiveness of our PTS, we conduct experiments to assess the impact of domain prompt number $N$ in Eq.~\ref{function:pooling} on SDG and multi-source DG setting, respectively, as shown in Fig.~\ref{pic:parameter_analysis_N}.
Even under the multi-source DG setting with more source domains, PTS can still improve the model's generalization performance as the number of $N$ increases.
To further demonstrate the stability of our PTS, we studied the impact of different initialization of the manually designed textual domains in $\mathcal{P}$, as shown in Tab.~\textcolor{red}{\ref{tab:initialization_analyse}}.
It is observable that our PTS shows slight fluctuations in individual benchmark datasets, demonstrating the insensitive to the initialization of $\mathcal{P}$.

\begin{figure}[!htbp]
\vspace{-2.0mm}
    \centering
    \includegraphics[width=0.47\textwidth]{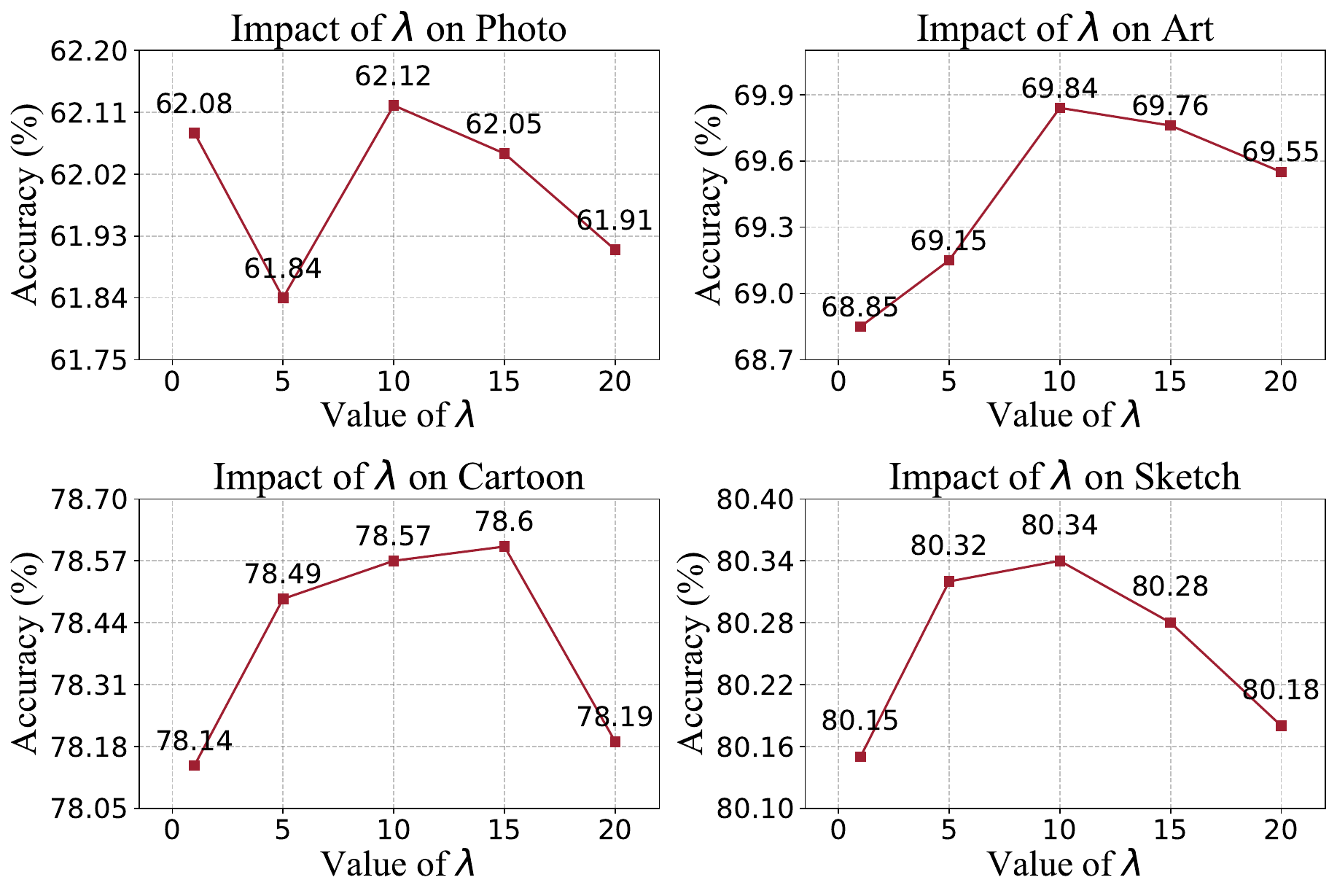}
    \vspace{-3.0mm}
    \caption{
    Parameter analysis of $\lambda$ in Eq.~\ref{function:reward_all} on PACS dataset.
    }
    \label{pic:parameter_analysis_lambda}
    \vspace{-5.5mm}
\end{figure} 
\subsection{Parameter Analysis}
We conduct experiments to analyze the impact of the hyper-parameters in our method on PACS dataset, including $\lambda$ in Eq.~\ref{function:reward_all}. The results are shown in Fig.~\ref{pic:parameter_analysis_lambda}.
$\lambda$ is a parameter to balance the trade-off between generating images in diverse domain styles and preserving key categorical features.
Based on Fig.~\ref{pic:parameter_analysis_lambda}, we set $\lambda=10$ for all datasets. 

\section{Conclusion} 
In this paper, we propose the Progressive Adversarial Domain Prompt Tuning framework based on the pre-trained T2I foundation model on SDG. 
Our method leverages adversarial training to learn two sets of abstract prompts, one that captures domain-invariant category information and another that models domain-specific styles.
Furthermore, we employ the Progressive Training Scheme to enhance the diversity of domain styles.
Extensive experiments on multiple DG datasets have demonstrated that our framework outperforms existing state-of-the-art SDG methods.
{
    \small
    \bibliographystyle{ieeenat_fullname}
    \bibliography{main}
}


\end{document}